\documentclass[runningheads]{llncs}
\usepackage[T1]{fontenc}
\usepackage{graphicx}
\usepackage{booktabs}
\usepackage{amsmath}
\usepackage{amssymb}
\usepackage{amsfonts}
\usepackage{bbold}
\usepackage{subcaption}
\usepackage{arydshln}
\usepackage{multirow}
\usepackage{hyperref}
\usepackage{tablefootnote}

\usepackage{mwe}

\begin{document}

\title{Deep Domain Isolation and Sample Clustered Federated Learning for semantic segmentation}
\toctitle{Deep Domain Isolation and Sample Clustered Federated Learning for semantic segmentation}

\titlerunning{Deep Domain Isolation and Sample Clustered Federated Learning}

\author{Matthis Manthe\inst{1,2} \orcidID{0009-0003-0425-880X}\and
Carole Lartizien\inst{2}\orcidID{0000-0001-7594-4231} \and
Stefan Duffner\inst{1}\orcidID{0000-0003-0374-3814}}
\tocauthor{Matthis Manthe, Carole Lartizien, Stefan Duffner}

\authorrunning{M. Manthe et al.}

\institute{Univ Lyon, INSA Lyon, CNRS, UCBL, Centrale Lyon, Univ Lyon 2, LIRIS, UMR5205, F-69621 Villeurbanne, France \email{\{matthis.manthe,stefan.duffner\}@insa-lyon.fr}
 \and
 Univ Lyon, CNRS, INSA Lyon, UCBL, Inserm, CREATIS UMR 5220, U1294, F‐69621 Villeurbanne, France \email{carole.lartizien@creatis.insa-lyon.fr}}

\maketitle              

\begin{abstract}
Empirical studies show that federated learning exhibits convergence issues in Non Independent and Identically Distributed (IID) setups. However, these studies only focus on label distribution shifts, or concept shifts (e.g. ambiguous tasks). In this paper, we explore for the first time the effect of covariate shifts between participants' data in 2D segmentation tasks, showing an impact way less serious than label shifts but still present on convergence. Moreover, current Personalized (PFL) and Clustered (CFL) Federated Learning methods intrinsically assume the homogeneity of the dataset of each participant and its consistency with future test samples by operating at the client level. We introduce a more general and realistic framework where each participant owns a mixture of multiple underlying feature domain distributions. To diagnose such pathological feature distributions affecting a model being trained in a federated fashion, we develop Deep Domain Isolation (DDI) to isolate image domains directly in the gradient space of the model. A federated Gaussian Mixture Model is fit to the sample gradients of each class, while the results are combined with spectral clustering on the server side to isolate decentralized sample-level domains. We leverage this clustering algorithm through a Sample Clustered Federated Learning (SCFL) framework, performing standard federated learning of several independent models, one for each decentralized image domain.
Finally, we train a classifier enabling to associate a test sample to its corresponding domain cluster at inference time, offering a final set of models that are agnostic to any assumptions on the test distribution of each participant. We validate our approach on a toy segmentation dataset as well as different partitionings of a combination of Cityscapes and GTA5 datasets using an EfficientVIT-B0 model, showing a significant performance gain compared to other approaches. Our code is available at \href{https://github.com/MatthisManthe/DDI_SCFL}{https://github.com/MatthisManthe/DDI\_SCFL}.

\end{abstract}

\section{Introduction}

Federated learning was initially proposed in \cite{mcmahan_communication_efficient_2017} as a decentralized privacy-preserving machine learning paradigm enabling multiple data owners to collaboratively train a model on their combined data without ever sharing them, introducing the now well established Federated Averging (FedAvg) algorithm. It was rapidly shown that this paradigm suffers from Non Independently and Identically Distributed (Non-IID) samples between clients \cite{zhao_federated_2018}, with a large amount of works devoted to solve this problem since then \cite{sahu_federated_2018,sattler_clustered_2019,li_ditto_2021,marfoq_federated_2021,wu_personalized_2023}. One can summarize the current state-of-the-art methods to mitigate the effect of Non-IID training data on convergence into three categories. First, \textit{global} methods, such as FedProx \cite{sahu_federated_2018} or SCAFFOLD \cite{karimireddy_scaffold_2020}, alter the server-side aggregation or regularize local learning to obtain one better single model at the end. Later, the notion of \textit{personalized} methods (PFL) emerged, such as FedPer \cite{arivazhagan_federated_2019}, pFedMe \cite{dinh_personalized_2020}, Ditto \cite{li_ditto_2021} or FedEM \cite{marfoq_federated_2021}. They focus on training one model per participant in the federation whilst still benefiting from collaboration. Finally, \textit{hybrid} or \textit{clustered} methods (CFL) were also proposed in case of existence of groups of clients with similar data distribution. We can cite the original CFL method \cite{sattler_clustered_2019}, or IFCA \cite{ghosh_efficient_2020}. 

Almost all state-of-the-art methods experimented on Non-IID setups with label shifts or concept shifts. Only few works examined the effect of covariate shift on federated learning in segmentation tasks such as FedDrive \cite{fantauzzo_feddrive_2022}, while using batch normalization layers. They show that SiloBN \cite{andreux_siloed_2020} enables to recover the lost performance from FedAvg. 
One question that remains unexplored is how covariate shifts can affect FedAvg for segmentation tasks when using normalization layers which do not require batch statistics updates, such as Instance Normalization \cite{ulyanov_instance_2017} more standard in segmentation models \cite{isensee_nnu_net_2021}, or Group and Layer Normalization as in more recent vision transformers \cite{dosovitskiy_image_2020}, which we try to answer in this article.

Moreover, personalized and clustered federated learning trade in a decrease of generalization of final models for a faster convergence and better local performance, i.e. desired specialization. Depending on the segmentation task and practical application, one can prefer personalizing or clustering models for one type of shift while preserving their generalization power for others. In this article, we explore this idea for covariate shifts, susceptible to be \textit{true underlying image domains}. In the case of medical image segmentation, different hospitals are likely to use scanners from a small set of existing ones potentially altering the appearance of the elements to segment. In autonomous driving, the features distribution of images obtained in a country (road, vegetation and people appearance, ambient luminosity and colours, cars' brands, etc.) is quite likely to be similar in neighbouring countries, while relatively constant in time. Thus, we propose to build a federated clustering algorithm focusing on types of covariate distributions (i.e. image domains). We resume federated learning on each decentralized cluster to perform focused clustered federated learning. For each cluster, global methods such as SCAFFOLD can be used to fix the convergence issues due to local label distribution discrepancy if any, preserving the generalization power of the clustered models with respect to labels and concepts.\\

Finally, current personalized and clustered federated methods intrinsically assume the homogeneity of the dataset of each participant and its consistency with future test samples by working at a client level. The existence of multiple underlying feature domains from which each participant owns a mixture of is a more general and realistic framework (large hospitals can own multiple scanners). This idea was already proposed in FedEM \cite{marfoq_federated_2021}, and extended to include covariate and concept shifts recently in \cite{wu_personalized_2023}, making these the closest works to ours. Both methods output a collection of models and a set of weights for a linear combination of their output per client. They thus remain client-based, assuming the strict consistency of future test samples of a client with its training ones, and requiring labeled samples for clients outside of the federation to compute its combination weights before the usage of the models. These are two large constraints we try to remove. We instead propose to perform the clustering of image domains at the sample level, effectively defining a Sample Clustered Federated Learning (SCFL) framework. We train a classifier after clustering, enabling to associate a test sample to its corresponding cluster at inference time. The set of cluster models combined with a cluster assignment method becomes self-sufficient on any test sample without any further assumptions.

\paragraph{Summary of contributions}
\begin{itemize}
    \item We explore for the first time the impact of covariate shifts on federated learning for 2D segmentation tasks without batch normalization layers. 
    \item We formalize a novel federated framework, named \textbf{Sample Clustered Federated Learning (SCFL)}, to resume federated training on multiple simplified decentralized image domains enhancing federated convergence speed. The final set of cluster models operate in a way that is agnostic to any assumptions on the further test samples of a client. It enables local test distribution shifts without loss in performance, and direct application of the set of models on clients outside of the federation.
    \item We develop \textbf{Deep Domain Isolation (DDI)}, a clustering method in class-specific gradient spaces of a model during federated training to isolate image domains \textit{orienting the optimization in different directions}. 
    It focuses its action on \textit{image domains requiring a model with different features to perfectly segment them}, in a centralized or decentralized fashion. This can call attention to a mismatch between segmentation task and model or, in the federated case of interest, a Non-IID distribution potentially altering convergence.
\end{itemize}

\section{Method}

\subsection{Notations, global federated objective and FedAvg}

\subsubsection{Data distribution and models} Let's assume the existence of $M$ image domains and $K$ clients, each with a local dataset $D_k:=\{(x_{k,i}, y_{k,i})\}_{i=1}^{n_k}$ composed of an arbitrary number of images from each domain, $x_{k,i}\in\mathcal{X} = \mathbb{R}^{3\times h\times d}$ the image to segment and $y_{k,i}\in\mathcal{Y} = \mathcal{C}^{h\times d}$ its ground-truth segmentation map, $\mathcal{C}=[C]$ the set of $C$ classes. We note $n_k$ the local dataset size of institution $k$ and $N = \sum_{k=1}^Kn_k$ the total number of samples. We note $w\in\mathcal{W} = \mathbb{R}^p$ the parameters of the neural network to be optimized for the downstream segmentation task.

\subsubsection{Global federated learning} Using $L:\mathcal{W}\times\mathcal{X}\times\mathcal{Y}\rightarrow \mathbb{R}$ a pixel-wise loss function 
\begin{equation}
    L(w,x,y)=\frac{1}{hd}\sum_i^h\sum_j^dl(w,x, y^{i,j})
\end{equation} such as a Cross-Entropy loss, the standard federated objective can be written
\begin{equation}
    w^* = \underset{w\in\mathcal{W}}{argmin } \sum_{k=1}^K\sum_{i=1}^{n_k}L(w, x_{k,i}, y_{k,i})\label{eq:global_fed_obj} \; .
\end{equation} With FedAvg \cite{mcmahan_communication_efficient_2017}, during each communication round $t$, each client performs $E$ local epoch(s) of Stochastic Gradient Descent (SGD) using the previous global model $w^t$ as initialization. These $K$ updates $\Delta w_k^t = w_k^t-w^t$ are communicated to the server and aggregated following $w^{t+1} = w^t + \sum_{k=1}^K\frac{n_k}{N}\Delta w_k^t$ to conclude a communication round.

\subsection{Sample Clustered Federated Learning (SCFL)}

\begin{figure}[h!]
    \centering
    \includegraphics[scale=0.23]{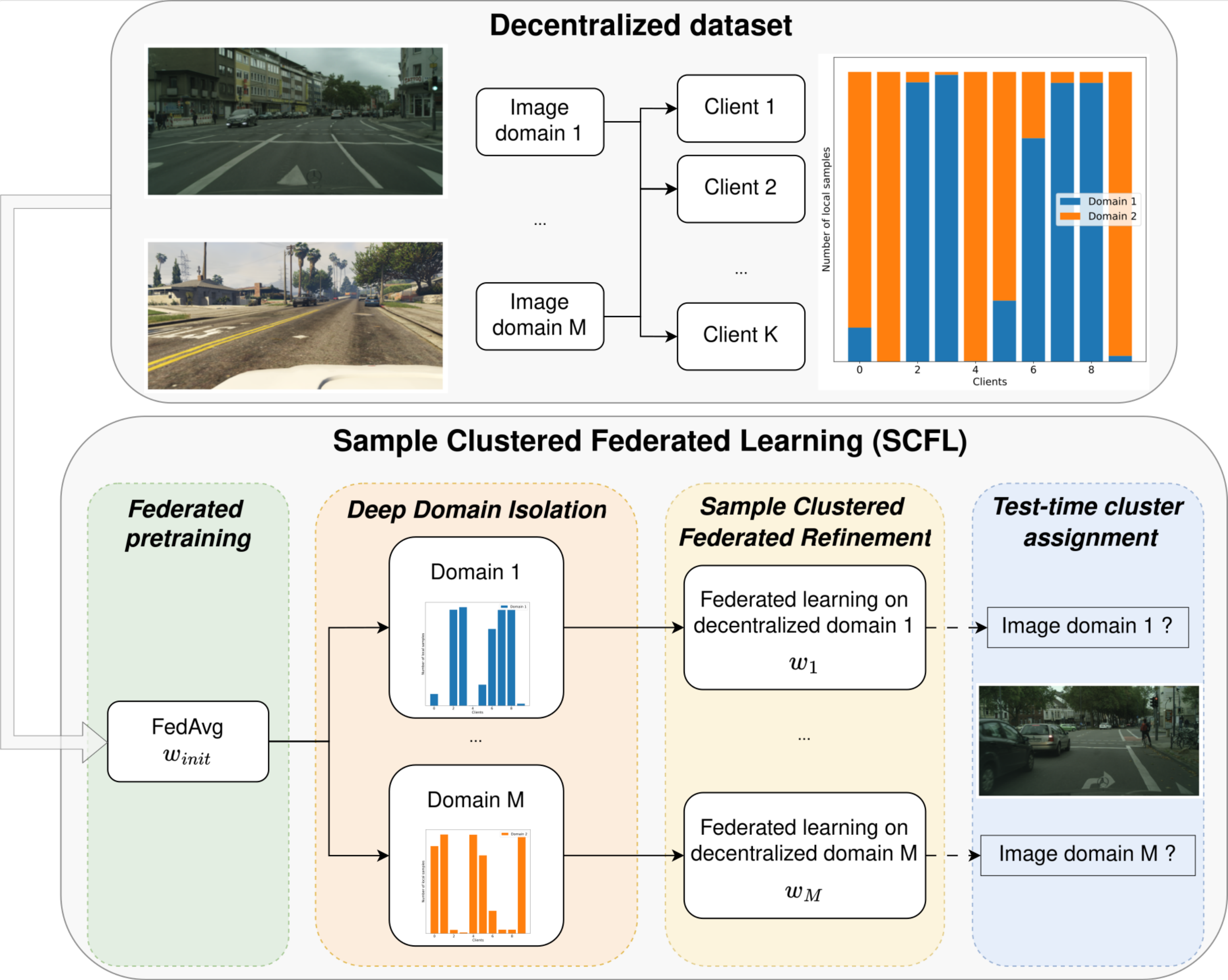}
    \caption{Overview of the Sample Clustered Federated Learning (SCFL) framework. It is composed of four sequential blocks. \textbf{1. Federated pretraining} providing a model with informative gradients and a good initialization for further domain specialization. \textbf{2. Deep Domain Isolation} clustering image domains in a federated fashion. \textbf{3. Sample Clustered Federated Refinement} performing federated learning on each decentralized image domain. \textbf{4. Test-time cluster assignment} on a test sample to select its matching cluster model.}
    \label{fig:Overview}
\end{figure}

We provide in Figure \ref{fig:Overview} an overview of the proposed framework. It is composed of four sequential blocks. 

\subsubsection{1. Federated Pretraining.} We first apply FedAvg for several communication rounds on the complete federation to obtain an initial model $w_{init}$. It enables the model to be in a state where sample gradients are informative. A model benefits from being trained conjointly on all image domains as long as they are not conflicting. This primary step gives a good initialization for a well-performing specialization on each domain.

\subsubsection{2. Clustering.} We then apply our Deep Domain Isolation, a federated gradient clustering method isolating image domains while ignoring sample label distributions. We extensively describe this method in Section \ref{sec:deep_dom_isol}. It provides a sample-level clustering $Cluster_{w_{init}}:\mathcal{X}\times\mathcal{Y} \rightarrow [M]$.

\subsubsection{3. Sample Clustered Federated Refinement.} Standard FedAvg on the complete federation leads to complicated local optimizations and can lead to client drift similarly to label or concept shifts in case of Non-IID federations
\cite{karimireddy_scaffold_2020}. We apply FedAvg on each isolated cluster using $w_{init}$ as the initialization, effectively solving
\begin{equation}
    \{w^*_m\}_{m=1}^{M} = \underset{\{w_m\}_{m=1}^{M}\in\mathcal{W}^M}{argmin } \sum_{k=1}^{K}\sum_{i=1}^{n_k}L(w_{Cluster_{w_{init}}(x_{k,i}, y_{k,i})}, x_{k,i}, y_{k,i})\label{eq:clustered_fed_obj}
\end{equation} with a common pretraining between image domains. This objective differs from standard CFL \cite{sattler_clustered_2019} as each client can contribute to multiple clusters if they own images from multiple domains. This produces one cluster model per domain. Note that a variety of global optimizers such as SCAFFOLD could be used instead.

\subsubsection{4. Test-time cluster assignment and inference.} To select the right cluster model for a label-free test image, we train a classifier using as input an image, and outputting a cluster identifier in a federated fashion to mimic the computed clustering function $Cluster_{w_{init}}$. Training details to make it converge in a Non-IID setup are provided in Section \ref{sec:source_classif_detail}.

\subsection{Deep Domain Isolation (DDI) for semantic segmentation}\label{sec:deep_dom_isol}

Given the parameters of a model $w_{init}$ and a decentralized dataset $D:=\bigcup_{k=1}^{K}D_k$, we aim at isolating image domains from $D$ which are altering the current optimization, in a federated fashion i.e. \textit{without ever sharing a large amount of information at sample-level to the server.} We also want to limit the influence of the class distribution of a sample on the computed clusters.\\ 

To this end, we define the class-masked averaged losses
\begin{equation}
     L_c(w,x,y) = \frac{1}{\sum_i^h\sum_j^d\mathbb{1}_{\{y^{i,j}=c}\}}\sum_i^h\sum_j^d\mathbb{1}_{\{y^{i,j}=c\}}l(w_{init},x,y^{i,j}), \forall c\in \mathcal{C} 
\end{equation} isolating the influence of a single class from the total loss. We note $D^c=\{(x_i,y_i) | \ i<N,\  \sum_i^h\sum_j^d\mathbb{1}_{\{y_{i,j}=c\}} > 0\}$ the decentralized subset of samples containing at least one pixel of a class $c$. The hyperspheres of normalized class-specific gradients to be clustered are $G_c = \{\frac{\nabla L_c(w_{init},x,y)}{||\nabla L_c(w_{init},x,y)||}|\ \{x,y\}\in D^c\}$.\\

We compute a Federated Gaussian Mixture model (Fed-GMM) proposed in \cite{wu_personalized_2023}  with $M$ clusters and diagonal covariance matrices on each set $G_c$. 
While other models such as von Mises-Fisher mixtures might have been more adapted to fit distributions on hyperspheres of normalized class-specific gradients, we chose GMMs with diagonal covariance matrices for their simplicity, straightforward applicability in a federated setup and a per-parameter variance estimation well-adapted to the exploration of gradient spaces, at the cost of representation power. This provides 
\begin{equation}
    \Pi_c\in\mathbb{R}^M, \ \ \mu_{c,m}\in\mathbb{R}^p, \ \ \Sigma_{c,m}\in\mathbb{R}^{p\times p}, \ \forall m < M
\end{equation} the optimized mixture weights, means and diagonal covariance matrices, giving membership likelihoods $s_{c,(x,y),m}$ of a sample $(x,y)$ to a cluster $m$ for a class $c$ following 
\begin{equation}
    \forall (x,y)\in D^c, \ s_{c,(x,y),m}\propto \Pi_{c,m}\mathcal{N}(\frac{\nabla L_c(w_{init},x,y)}{||\nabla L_c(w_{init},x,y)||} | \mu_{c,m}, \Sigma_{c,m})
\end{equation} and class-wise normalized membership vectors $\hat{s}_{c,(x,y)} := \frac{[s_{c,(x,y),0},\ ...,\ s_{c,(x,y),M}]}{\sum_{m=1}^Ms_{c,(x,y),m}}$.

These membership vectors are communicated to the server and used to compute a similarity measure between samples server-side. With $\mathcal{C}_{y_i,y_j} = \{c \ |\ c\in\mathcal{C}, \ (x_i,y_i)\in D^c, \ (x_j,y_j)\in D^c\}$ the common classes between two train samples $(x_i,y_i)$ and $(x_j,y_j)$, we define their similarity $\hat{S}$ as the average similarity between their normalized cluster assignments for each of their common classes
\begin{equation}
    \hat{S}_{(x_i,y_i), (x_j,y_j)} = \frac{1}{|\mathcal{C}_{y_i,y_j}|}\sum_c^{\mathcal{C}_{y_i,y_j}} S(\hat{s}_{c,(x_i,y_i)}, \hat{s}_{c,(x_j,y_j)})
\end{equation} based on a chosen similarity function $S$ between two discrete distributions with $M$ values. We used the Bhattacharyya coefficient defined as 
\begin{equation}
    \forall p_i,p_j\in[0,1]^M, S(p_i,p_j) = \sum_m^M\sqrt{p_{i,m}p_{j,m}}
\end{equation}

This finally gives a similarity matrix $\hat{\mathcal{S}} = [\hat{S}_{(x_i,y_i), (x_j,y_j)}]_{i,j < N}$ between each pair of samples server-side, on which we perform spectral clustering to obtain a sample-level decentralized clustering of image domains, for which the number of final cluster $M$ is a hyperparameter. Intuitively, sample class-specific gradients in segmentation are sufficiently well behaved and averaged to enable simple diagonal GMMs to work decently, while we use the presence of multiple classes on each sample to link class-specific clusterings and compute a final sample-level clustering. This would not generalize to single-label classification tasks.

\subsubsection{Complexity reduction} This process can be expensive in memory (saving multiple gradients locally), time (computations on large sized-gradients) and communication (GMMs parameters represent $2Mp$ values). Since the effect of covariate shift tends to be redundant on the large amount of parameters of neural networks, the server can send to the participant a random list of parameter indices, to then perform Deep Domain Isolation on pruned gradients. We name this variant Pruned Deep Domain Isolation, with a parameter $\alpha \in [0,1]$ defining the proportion of saved parameters in gradients. With $\alpha$ sufficiently small, the cost in time, memory and communication of this clustering becomes negligible compared to federated training, while potentially helping Federated GMMs to converge better by decreasing the dimensionality of the problem.

\section{Experiments}
\subsection{Datasets and models}

Main experiments were led on two segmentation tasks. While we propose the SCFL framework to work on $M$ image domains, we experimented on datasets composed of two domains each, enabling a better control on the introduced covariate shift while simplifying the limitation of other forms of shifts.

\subsubsection{TMNIST-Inv}
We generated this segmentation toy dataset using MNIST, named Triple MNIST Segmentation (TMNIST-Seg). An example is given in Figure \ref{fig:clear}. Each $64\times 96$ image is composed of three randomly selected 0, 1, 3 and 4 digits from MNIST. The ground-truth segmentation masks are 0 intensity pixels for the background class, and non-zero pixels of class 1 to 4 for each digit 0, 1, 3 and 4 in order, giving a 5 classes segmentation task. There are 50 samples for each arrangement of three digits, giving 3200 samples in training with a completely homogeneous label distribution. 1280 validation and test samples were generated the same way, with train and validation samples using train samples from MNIST, and test images using test samples from MNIST. 

\begin{figure}[t]
     \centering
     \begin{subfigure}[b]{0.6\textwidth}
         \centering
         \includegraphics[scale=0.2]{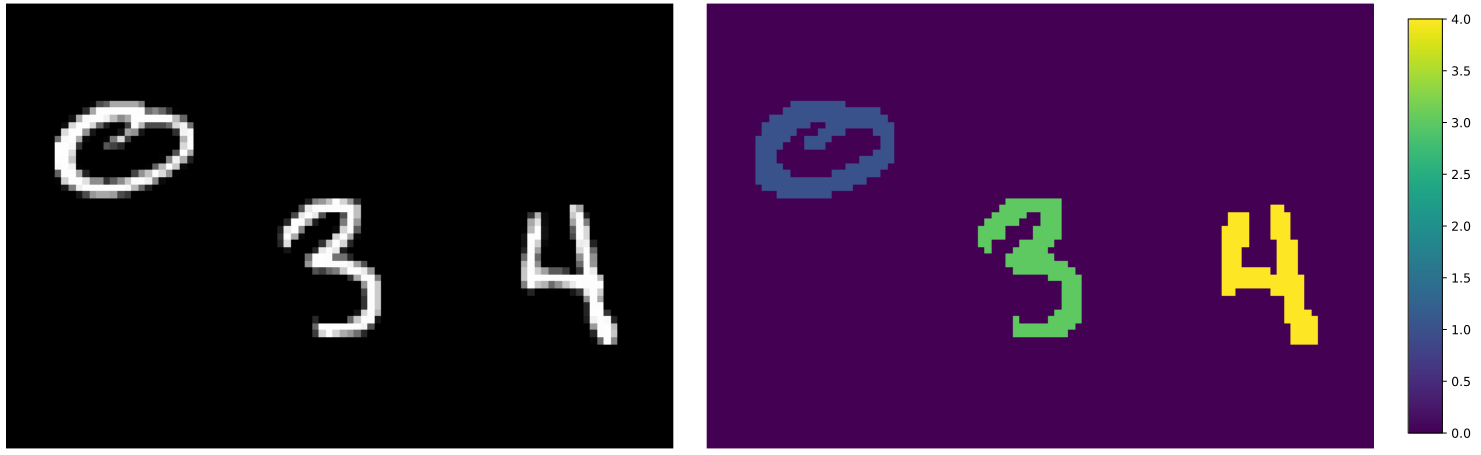}
         \caption{Clear image (left) and labels (right)}
         \label{fig:clear}
     \end{subfigure}
     \hfill
     \begin{subfigure}[b]{0.3\textwidth}
         \centering
         \includegraphics[scale=0.2]{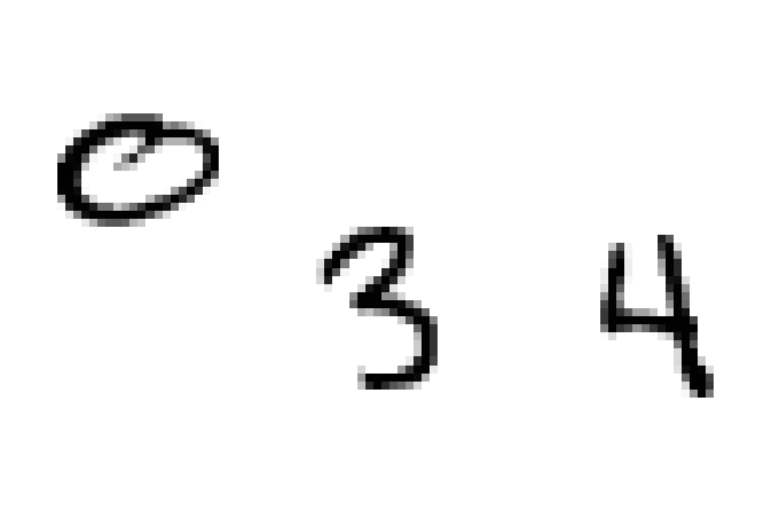}
         \caption{Grayscale inverted}
         \label{fig:inv}
     \end{subfigure}
        \caption{TMNIST-Inv example}
        \label{fig:TMNIST_seg}
\end{figure}

To simulate a strong covariate shift, we applied a grayscale inversion (Figure \ref{fig:inv}) to half of this clean dataset (on train, validation and test sets). We name the resulting dataset TMNIST-Inv. We chose a simplification of a UNet-type segmentation model of around 71k parameters using instance normalization layers and reducing the decoder to one full resolution convolutional layer followed by a point-wise convolution for this task.

\subsubsection{Cityscapes+GTA5}
\begin{figure}[t]
     \centering
     \begin{subfigure}[b]{1.\textwidth}
         \centering
         \includegraphics[scale=0.2]{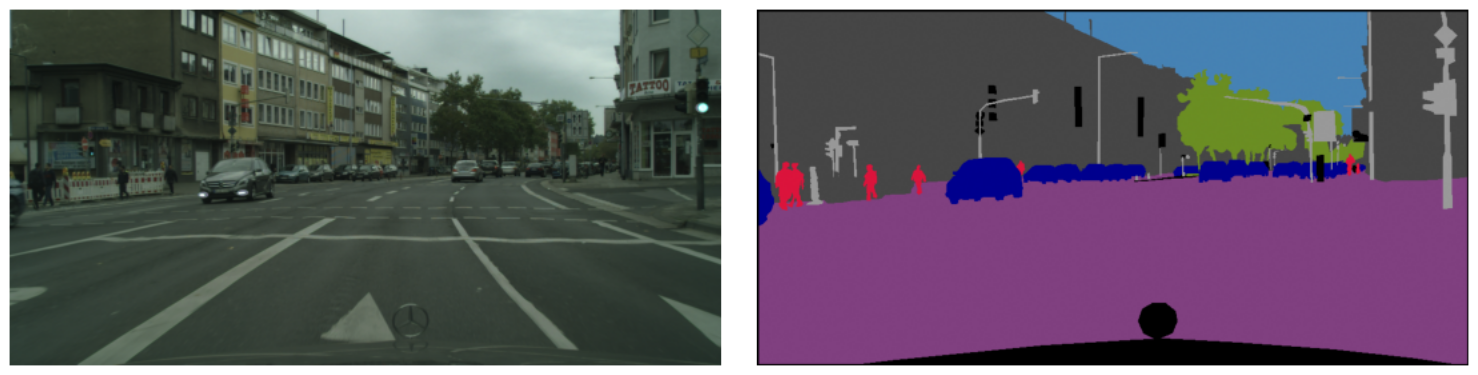}
         \caption{A sample from Cityscapes dataset.}
         \label{fig:City}
     \end{subfigure}
     \hfill
     \begin{subfigure}[b]{1.\textwidth}
         \centering
         \includegraphics[scale=0.2]{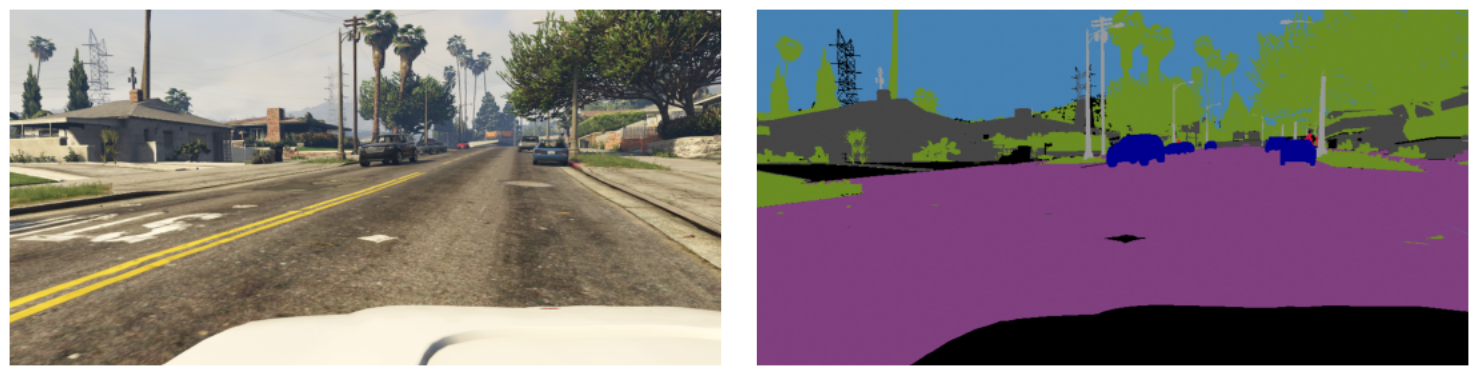}
         \caption{A sample from GTA5 dataset.}
         \label{fig:GTA}
     \end{subfigure}
        \caption{Two samples (image and target categories) from the two image domains of Cityscapes+GTA5 dataset.}
        \label{fig:CityGTA}
\end{figure}

We also use a more complex segmentation dataset from the autonomous driving field composed of the complete Cityscapes dataset \cite{cordts_cityscapes_2016} as well as randomly selected samples from the GTA5 dataset \cite{richter_playing_2016}. Cityscapes is composed of real photos of roads in Germany, while GTA5 consists of segmented images from the video game GTA5 (Figure \ref{fig:CityGTA}). While extremely consistent across classes, the domain shift present between these datasets is complex and currently studied as one of the state-of-the-art domain adaptation task in computer vision.

Three cities were isolated from Cityscapes train dataset to form the validation set, giving 5036 train, 914 validation and 1000 test samples, each composed of 50\% of each domain. To limit training times, every image (train and test) was resized to $512\times 1024$ pixels, using categories as the training targets and ignoring the background, giving a simplified 7-class segmentation task. We chose to use the EfficientVIT-B0 from \cite{cai_efficientvit_2023} for its small number of parameters (around 700k) and competitive performance.

\subsection{Federated splits}
\begin{figure}[t]
    \centering
     \begin{subfigure}[b]{0.32\textwidth}
         \centering
         \includegraphics[scale=0.12]{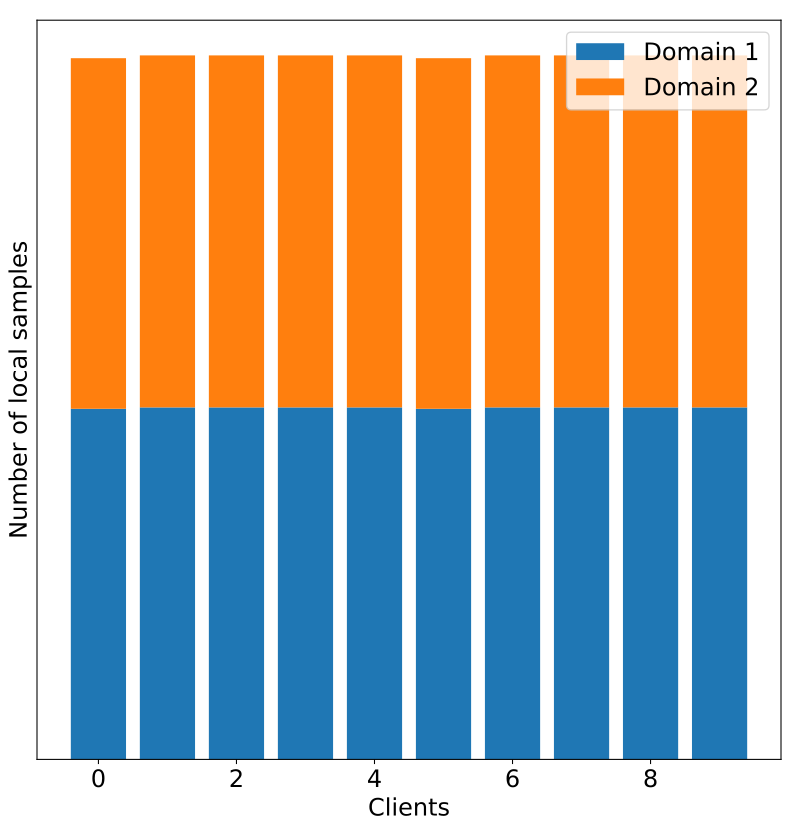}
         \caption{\textit{IID}}
         \label{fig:clear}
     \end{subfigure}
     \hfill
     \begin{subfigure}[b]{0.32\textwidth}
         \centering
         \includegraphics[scale=0.12]{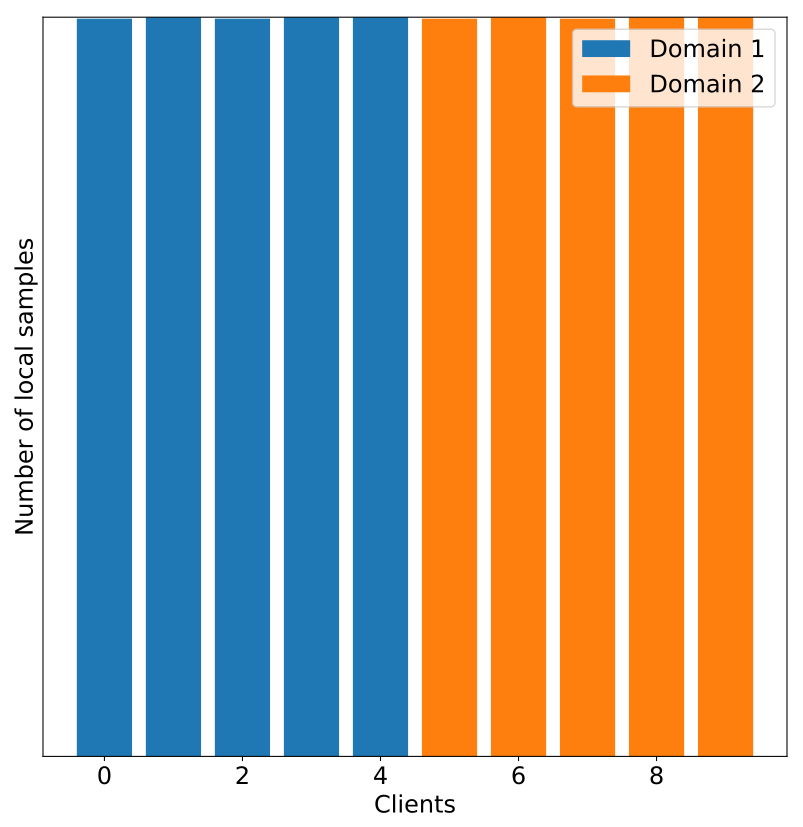}
         \caption{\textit{Full non-IID}}
         \label{fig:inv}
     \end{subfigure}
     \hfill
     \begin{subfigure}[b]{0.32\textwidth}
         \centering
         \includegraphics[scale=0.12]{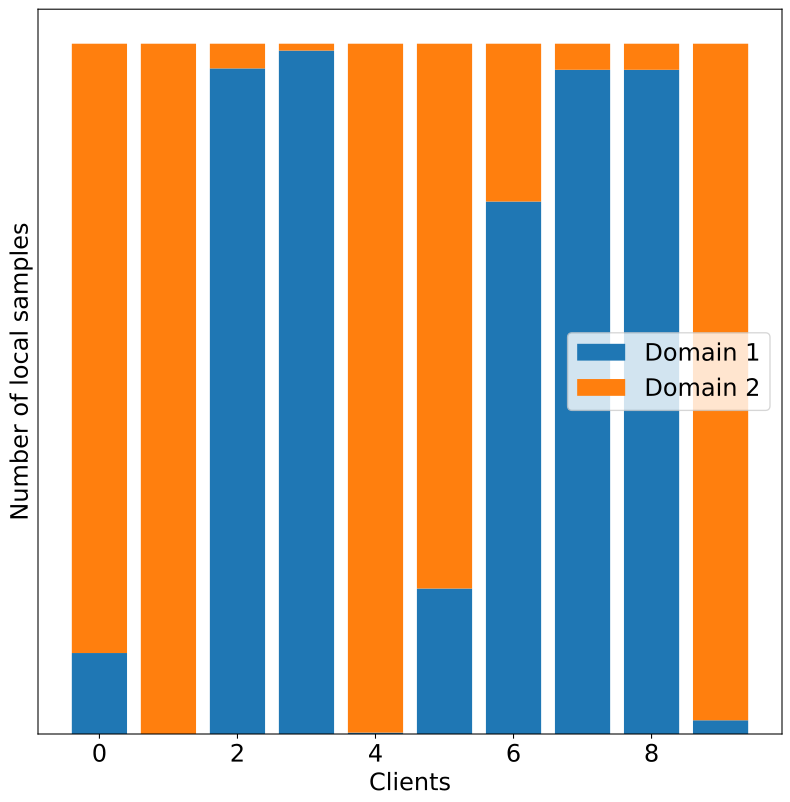}
         \caption{\textit{Dirichlet non-IID}}
         \label{fig:inv}
     \end{subfigure}
    \caption{Federated repartition of samples per client}
    \label{fig:distrib}
\end{figure}

For each dataset, we explore three data splits into 10 clients. An \textit{IID} split, with each client owning 50\% of their data from each domain. A \textit{Full non-IID} split, with each client owing samples from only one domain. Finally, a more realistic \textit{Dirichlet non-IID} partitioning sampled using a Dirichlet distribution with parameter 0.25. While there are no real clusters of participant In this last case, there remains two underlying image domains. These distributions are illustrated in Figure \ref{fig:distrib}.

\subsection{Baselines and experiments}
In Section \ref{sec:result_covariate_shift}, we first study \textbf{the effect of covariate shift on FedAvg} on models deprived of batch normalization layers using different local learning rates. In Section \ref{sec:results_full}, \textbf{we compare our approach to multiple baselines} on TMNIST-Inv and Cityscapes+GTA5 for each distribution. Included baselines are standard FedAvg \cite{mcmahan_communication_efficient_2017}, a state-of-the-art global method SCAFFOLD \cite{karimireddy_scaffold_2020}, a standard baseline of personalization local finetuning (FedAvg+), and standard clustered federated learning method CFL \cite{sattler_clustered_2019}. We also define Prior SCFL as applying SCFL with prior clustering and test-time cluster assignment as if the information of image domains was public. This mimics the "best possible" performance of SCFL for each dataset and split at the cost of privacy. To illustrate the limitations of Personalized and Clustered FL, we also test these methods trained on non-IID splits on an IID test split, simulating a distribution shift between local training and tests samples (named \textit{w/ test shift}). In Section \ref{sec:results_cluster}, we explore for each dataset \textbf{the clustering performance of our Deep Domain Isolation}, with and without random gradient pruning, and varying the chosen amount of communication rounds of FedAvg pretraining. Finally in Section \ref{sec:results_source_classif}, we explore for each dataset \textbf{the performance of the source domain classifier} on prior and computed imperfect clustering results. We report
semantic segmentation performance using the standard intersection over union metric averaged over all classes (mIoU). IoU computes the ratio of the intersection over the union of the predicted and ground truth segmentation for each class.

\subsection{Training parameters}

\subsubsection{Overall training setup} We use SGD locally with a batch size of 16 for the toy dataset and 2 for Cityscapes+GTA5. Federated optimizations were performed for a total of 700 communication rounds for every method, with one local epoch per round. If a split in multiple clusters was performed by CFL or SCFL at round T, 700-T rounds of clustered federated training was then performed. This does not enable to reach complete convergence for Cityscapes+GTA5, the required number of communication rounds is outside of our computational capabilities. Our experiments give however a clear idea of the convergence speed of each method. A dropout of 0.1 was applied after depth-wise convolution layers of MBConvs in the EfficientVIT-B0. We did not use any data augmentation. An exponential local learning rate decay of 0.9985 and 0.9975 was applied after each communication round for Cityscapes+GTA5 and TMNIST-Inv respectively. Local finetuning (FedAvg+) was performed on the best FedAvg model after 600 rounds of training and selected on best local validation for 100 local epochs. As the hyperparameters of the original CFL method \cite{sattler_clustered_2019} were impossible to tune, the communication round to which the federation was split in two is a hyperparameter for CFL and SCFL. Best round splits were 30 for TMNIST-Inv with IID and Dirichlet non-IID splits and 100 for the Full non-IID. It was 500, 400 and 500 in Cityscapes+GTA5 for the IID, Full non-IID and Dirichlet non-IID splits respectively. Implementations were made in Pytorch and computations performed on one NVIDIA V100 GPU and 80GB of RAM. Training one model on the Cityscapes+GTA5 dataset takes around 30 hours.

\subsubsection{SCFL's Source classifier training}\label{sec:source_classif_detail} In an IID federated split, training a domain classifier is trivial. In the pathological Full non-IID setup however, training such a classifier is a classification task where each client owns samples of only one class. They can also own mislabeled samples as our clustering is not perfect. Applying FedAvg on such a task does not converge for both datasets. We used SCAFFOLD \cite{karimireddy_scaffold_2020} instead of FedAvg and a local weight decay of 0.001. We also used an extremely small local learning rate, 0.005, essentially training the CNN in a distributed full batch mode with SCAFFOLD's momentum. This was enough for the classifiers to converge to close to perfect accuracy in less than 100 communication rounds in every setup. 

\section{Results}

\subsection{On the effect of covariate shift on FedAvg}\label{sec:result_covariate_shift}

\begin{table}[t]
    \centering
    \caption{FedAvg test performance (mIoU) for different datasets, federated distributions and initial local learning rates $\lambda_0$. Performances were averaged over 3 independent training runs for TMNIST-Inv dataset.}
    \begin{tabular}{|c||c|c|c||c|c|c||c|c|c|}
        \hline
        \multirow{2}{*}{\textbf{Dataset}} & \multicolumn{3}{c||}{\textbf{TMNIST-Inv}} & \multicolumn{3}{c||}{\multirow{2}{*}{\textbf{TMNIST-Inv}}} & \multicolumn{3}{c|}{\multirow{2}{*}{\textbf{Cityscapes+GTA5}}} \\
        & \multicolumn{3}{c||}{(batch normalization)} & \multicolumn{3}{c||}{} & \multicolumn{3}{c|}{}\\\hdashline
        \multirow{2}{*}{\textit{Distribution}} & \multirow{2}{*}{\textit{IID}} & \textit{Full} & \textit{Dirichlet} & \multirow{2}{*}{\textit{IID}} & \textit{Full} & \textit{Dirichlet} & \multirow{2}{*}{\textit{IID}} & \textit{Full} & \textit{Dirichlet} \\
        & & \textit{non-IID} & \textit{non-IID} & & \textit{non-IID} & \textit{non-IID} & & \textit{non-IID} & \textit{non-IID}\\\hline\hline
        
        FedAvg & \multirow{2}{*}{\textit{0.941}} & \multirow{2}{*}{0.517} & \multirow{2}{*}{\textit{0.931}} & \multirow{2}{*}{0.866} & \multirow{2}{*}{0.865} & \multirow{2}{*}{0.871} & \multirow{2}{*}{0.731} & \multirow{2}{*}{0.728} & \multirow{2}{*}{0.735} \\
        $\lambda_0=0.032$ & & & & & & & & & \\\hline
        
        FedAvg & \multirow{2}{*}{\textit{0.940}} & \multirow{2}{*}{0.643} & \multirow{2}{*}{\textit{0.942}} & \multirow{2}{*}{0.894} & \multirow{2}{*}{0.899} & \multirow{2}{*}{0.898} & \multirow{2}{*}{0.753} & \multirow{2}{*}{0.749} & \multirow{2}{*}{0.756} \\
        $\lambda_0=0.1$ & & & & & & & & & \\\hline
        
        FedAvg & \multirow{2}{*}{\textit{0.936}} & \multirow{2}{*}{0.412} & \multirow{2}{*}{\textit{0.937}} & \multirow{2}{*}{\textit{0.933}} & \multirow{2}{*}{\textit{0.933}} & \multirow{2}{*}{0.923} & \multirow{2}{*}{0.762} & \multirow{2}{*}{0.764} & \multirow{2}{*}{\textbf{0.773}} \\
        $\lambda_0=0.32$ & & & & & & & & & \\\hline
    \end{tabular}
    \label{tab:covraiate_fedavg}
\end{table}

We show in Table \ref{tab:covraiate_fedavg} the performance of FedAvg on the chosen tasks. A model trained with FedAvg on TMNIST-Inv with batch normalization layers is highly affected by domain distribution, with \textbf{validation and test} performances (not training) collapsing in the Full non-IID case. For TMNIST-Inv, instance normalization and the low number of local iterations between aggregations seem enough to completely counter the effect of heterogeneous covariate shift on FedAvg whatever the learning rate and federated distribution. On Cityscapes+GTA5, a large initial learning rate tends to limit the gap between optimizations on IID and Non-IID splits, although only when using an abnormally high initial learning rate of 0.32. The local instability brought by such a high learning rate might mitigate local overfitting (e.g. client drift), surprisingly giving the best performances. Also, the Dirichlet non-IID distribution gives the best results when dealing with covariate shifts on Cityscapes+GTA5. Overall, with adapted normalization layers, the effect of covariate shift on FedAvg is very limited compared to what could be seen in the literature in case of label skew.

\subsection{SCFL performance and comparison to baselines}\label{sec:results_full}

\begin{table}[h]
    \centering
    \caption{Test performance (mIoU) of different methods on TMNIST-Inv and Cityscapes+GTA5. mIoUs were averaged over 3 independent training runs for TMNIST-Inv dataset.}
    \begin{tabular}{|c||c|c|c||c|c|c|}
        \hline
        \textbf{Dataset} & \multicolumn{3}{c||}{\textbf{TMNIST-Inv}} & \multicolumn{3}{c|}{\textbf{Cityscapes+GTA5}} \\\hdashline
        \multirow{2}{*}{\textit{Distribution}} & \multirow{2}{*}{\textit{IID}} & \textit{Full} & \textit{Dirichlet} & \multirow{2}{*}{\textit{IID}} & \textit{Full} & \textit{Dirichlet} \\
        & & \textit{non-IID} & \textit{non-IID} & & \textit{non-IID} & \textit{non-IID}\\\hline\hline
        FedAvg & 0.933 & 0.933 & 0.923 & 0.762 & 0.764 & 0.773 \\\hline
        SCAFFOLD & 0.930 & 0.916 & 0.913 & 0.763 & 0.763 & 0.756\tablefootnote{SCAFFOLD diverged on Cityscapes+GTA5 Dirichlet non-IID with a high learning rate. An initial learning rate of 0.1 instead of 0.32 was used in this case, justifying the poor result.}\\\hline
        FedAvg+ & \textbf{0.942} & 0.949 & 0.941 & 0.760 & 0.769 & 0.770 \\\hline
        CFL  & - & \textbf{0.970} & - & - & \textbf{0.776} & - \\\hline\hline
        Prior SCFL (ours) & 0.937 & \textbf{0.970} & \textbf{0.956} & 0.765 & \textbf{0.776} & \textbf{0.780} \\\hline
        SCFL (ours) & 0.937 & \textbf{0.970} & \textbf{0.956} & 0.764 & \textbf{0.775} & \textbf{0.780} \\\hline\hline
        FedAvg+ w/ test-shift & - & 0.866 & 0.925 & - & 0.754 & 0.768 \\\hline
        CFL w/ test-shift & - & 0.646 & - & - & 0.711 & - \\\hline
        SCFL w/ test-shift & - & \textbf{0.970} & \textbf{0.956} & - & \textbf{0.775} & \textbf{0.780} \\\hline
    \end{tabular}
    \label{tab:full_res}
\end{table}

We compare in Table \ref{tab:full_res} SCFL to other baselines. Our approach outperforms them in each tested case, with a clear performance gain in the Full and Dirichlet non-IID setups. Note that using SCFL on IID setups implies that the model of each cluster is trained using half the number of local iterations per communication round of standard FedAvg. The fact that a very slight gain in performance is perceptible in this case for both datasets is surprising.

For Full non-IID setups, SCFL and CFL are equivalent if the clusterings agree. They both bring a clear performance gain. CFL's clustering recovers the underlying image domains whatever the local learning rate for Cityscapes+GTA5, and only when using a small learning rate on TMNIST-Inv, while becoming highly inconsistent with the large learning rate enabling to reach the best federated performances. It is highly dependent on the local optimization scheme and affected by the symmetries of the loss landscape. Deep Domain Isolation does not have these limitations. We chose to put CFL in its best position by using a small learning rate during the rounds computing its clustering.

SCFL brings a performance gain over other baselines in the Dirichlet non-IID case when CFL is not applicable. 

Finally, evaluating personalized and clustered baseline methods trained on Full and Dirichlet non-IID splits on an IID test split (w/ test-shift entries) show their high vulnerability to test set distribution shifts. Since SCFL works at a sample level, it is not affected by this.

\subsection{Clustering performance}\label{sec:results_cluster}

\subsubsection{Deep Domain Isolation performance}

\begin{table}[t]
    \centering
    \caption{Clustering performance (rand index) of Deep Domain Isolation (DDI) with clustering performed at different split rounds during FedAvg training and with different domain distributions (seed 1-seed 2). A split round of 0 designates the clustering on a randomly initialized model. Pruned DDI was performed on gradients \textbf{after random selection of 1\% of the parameters by the server.}}
    \begin{tabular}{|c||c|c|c|c|c|c|}
        \hline
        & \textbf{Split round} & \textbf{0} & \textbf{300} & \textbf{400} & \textbf{500} & \textbf{600}\\\hline\hline
        
        \multirow{3}{*}{\textbf{DDI}} & \textit{IID} & \multirow{3}{*}{0.591} & \multicolumn{4}{c|}{0.9996 (1 mistake)} \\\cline{2-2}\cline{4-7}
        & \textit{Full Non-IID} & & 0.987-0.985 & 0.988-0.991 & 0.994-0.996 & 0.993-0.993 \\\cline{2-2}\cline{4-7}
        & \textit{Dirichlet non-IID} & & 0.987-0.993 & 0.992-0.994 & 0.996-0.996 & 0.994-0.994\\\hline\hline
        
        \multirow{3}{*}{\textbf{Pruned DDI}} & \textit{IID} & \multirow{3}{*}{-} & 0.998-0.997 & \multicolumn{3}{c|}{0.9996} \\\cline{2-2}\cline{4-7}
        & \textit{Full Non-IID} & & 0.986-0.987 & 0.982-0.989 & 0.983-0.992 & 0.993-0.990 \\\cline{2-2}\cline{4-7}
        & \textit{Dirichlet non-IID} & & 0.991 & 0.995-0.993 & 0.997-0.995 & 0.995\\\hline 
    \end{tabular}
    \label{tab:clustering_city}
\end{table}

A perfect clustering was obtained for TMNIST-Inv for any distribution and split round (rand index of 1.0). We show in Table \ref{tab:clustering_city} the agreement between Deep Domain Isolation results and image domains on models pretrained with FedAvg on Cityscapes+GTA5 for different number of communication rounds and federated splits for two seeds. It isolates close to perfectly the image domains in each setup. The more pretraining rounds are performed the better the clustering results, FedAvg converging towards an optima where image domains' gradients diverge into clusters. It also performs better in the IID setup than others, the federated distribution seems to alter the distribution of the gradients. 

\begin{figure}[t]
     \centering
     \begin{subfigure}[b]{0.47\textwidth}
         \centering
         \includegraphics[scale=0.1]{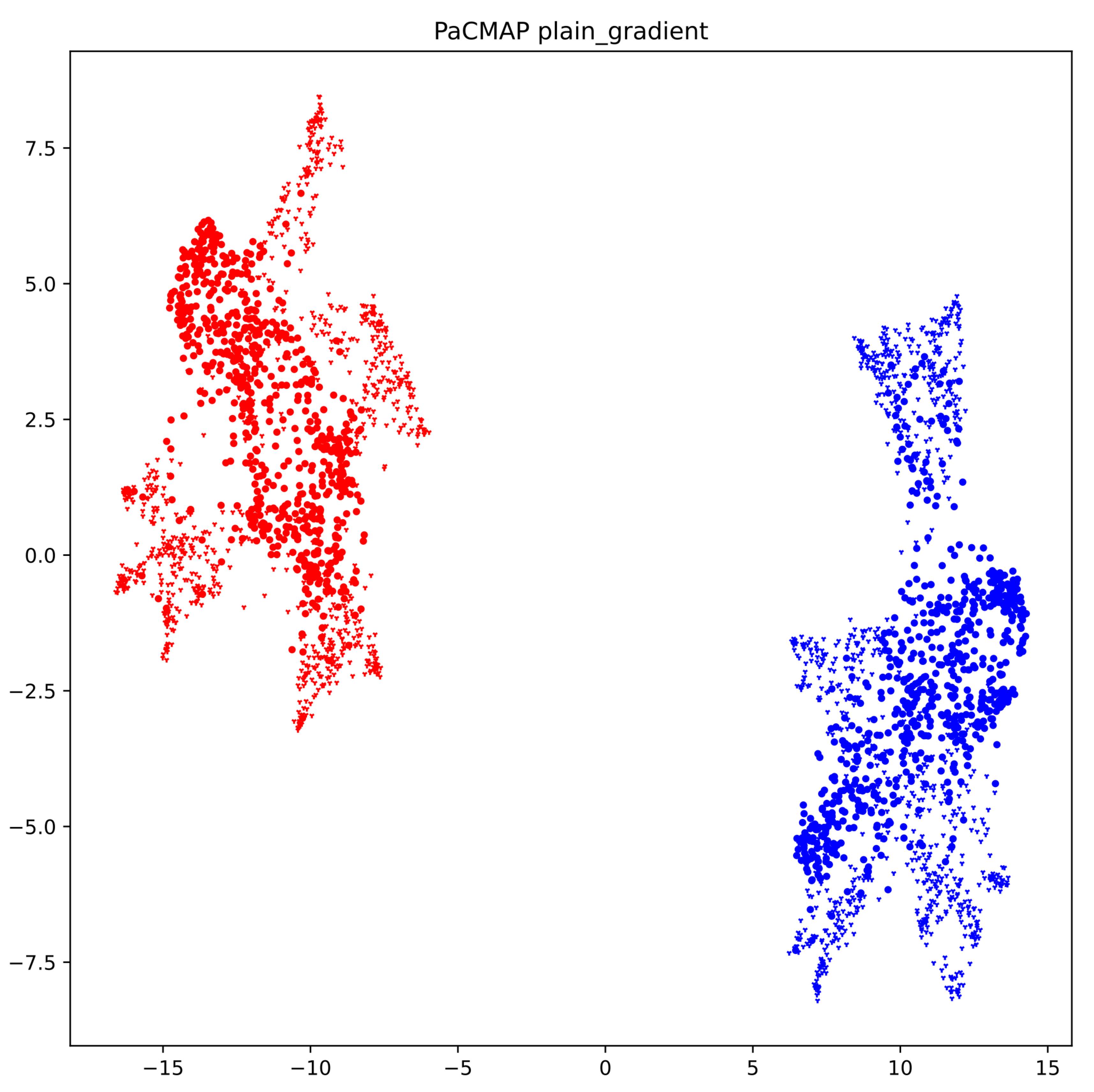}
         \caption{\textit{Plain} sample gradients (rand index 0.5)}
         \label{fig:GTA}
     \end{subfigure}
     \hfill
     \begin{subfigure}[b]{0.47\textwidth}
         \centering
         \includegraphics[scale=0.1]{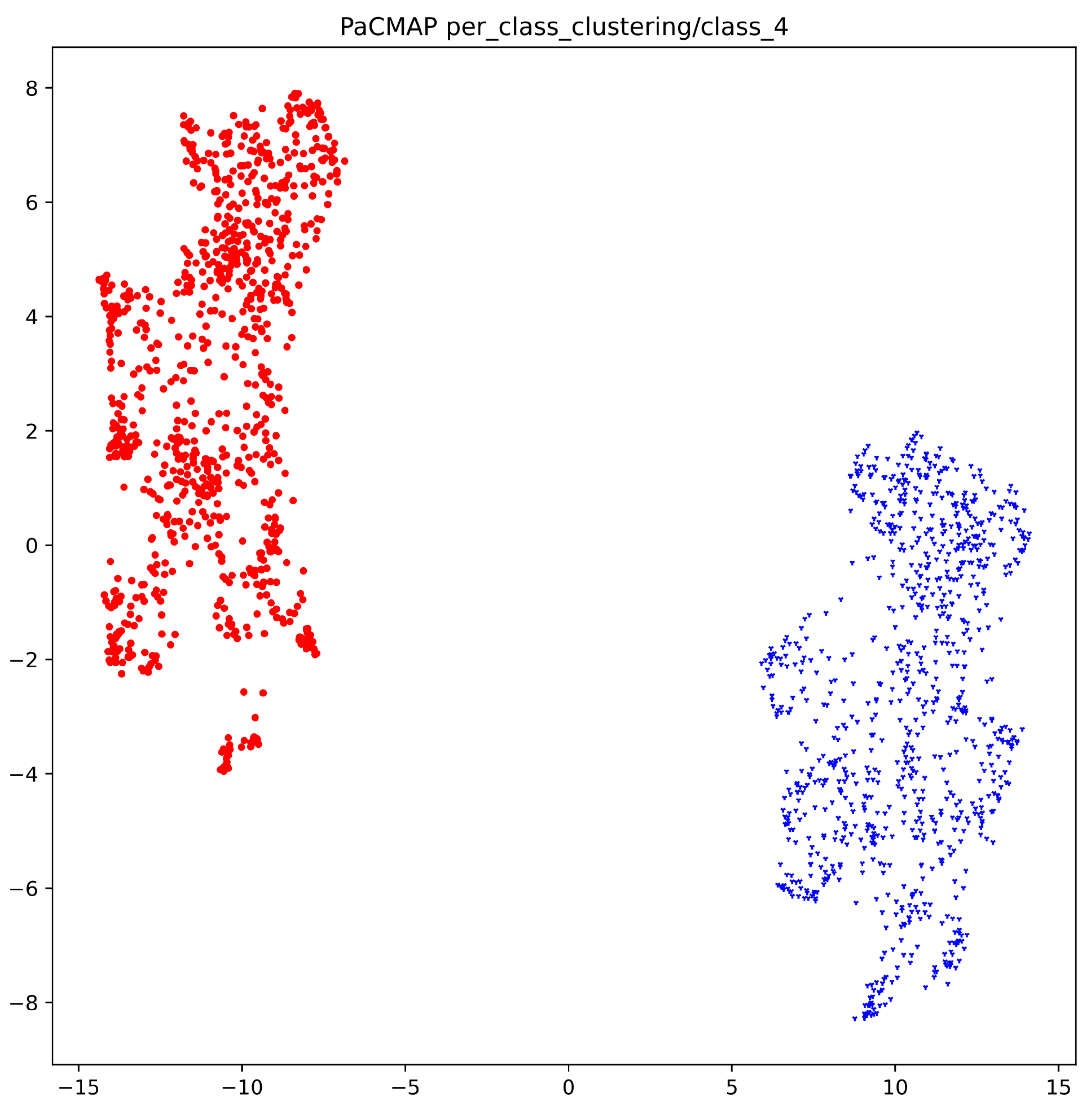}
         \caption{\textit{Digit 4} sample gradients (rand index 1.0)}
         \label{fig:GTA}
     \end{subfigure}
        \caption{PaCMAP representations of sample gradients of a model trained with FedAvg on TMNIST-Inv for 30 rounds on the Dirichlet non-IID split. Colors represent the image domains and markers the assigned clusters by Fed-GMM. Plain gradient clustering tends to be more difficult than per-class clustering. Note that PaCMAPs seem to nicely capture the image domains in both cases, while they are not applicable in a federated setup due to privacy constraints.}
        \label{fig:PaCMAP_TMNIST-Inv}
\end{figure}

We show in Figure \ref{fig:PaCMAP_TMNIST-Inv} PaCMAP representations of sample gradients, with colors representing the image domains and markers the assigned cluster by Deep Domain Isolation. Plain gradient clustering can be difficult compared to class-specific Fed-GMMs as it is affected by the class distribution of each sample. Class-specific gradients tend to behave in a simpler way.

\begin{figure}[t]
     \centering
     \begin{subfigure}[b]{0.47\textwidth}
         \centering
         \includegraphics[scale=0.1]{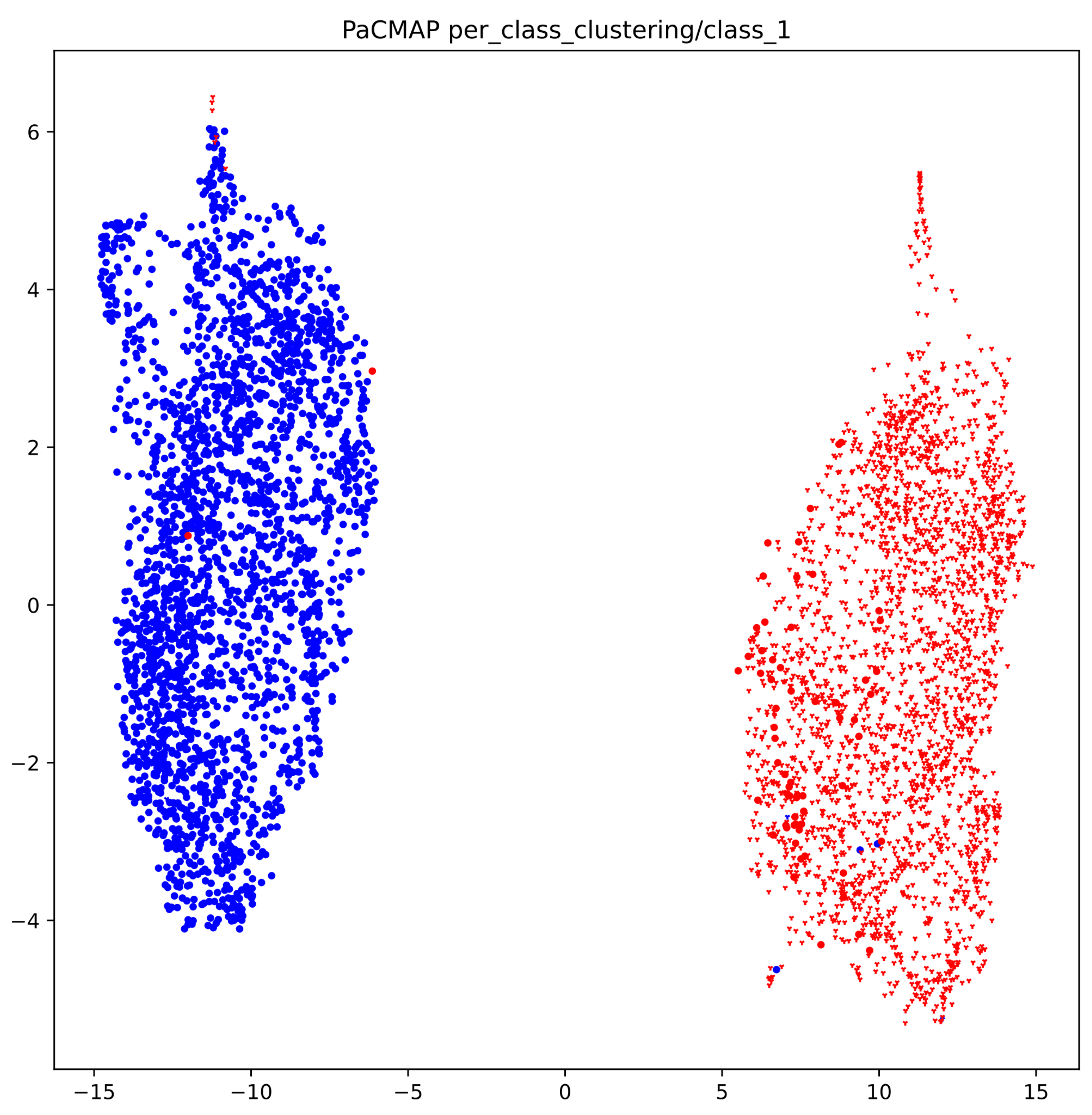}
         \caption{\textit{Class construction} sample gradients (rand index 0.97)}
         \label{fig:City}
     \end{subfigure}
     \hfill
     \begin{subfigure}[b]{0.47\textwidth}
         \centering
         \includegraphics[scale=0.1]{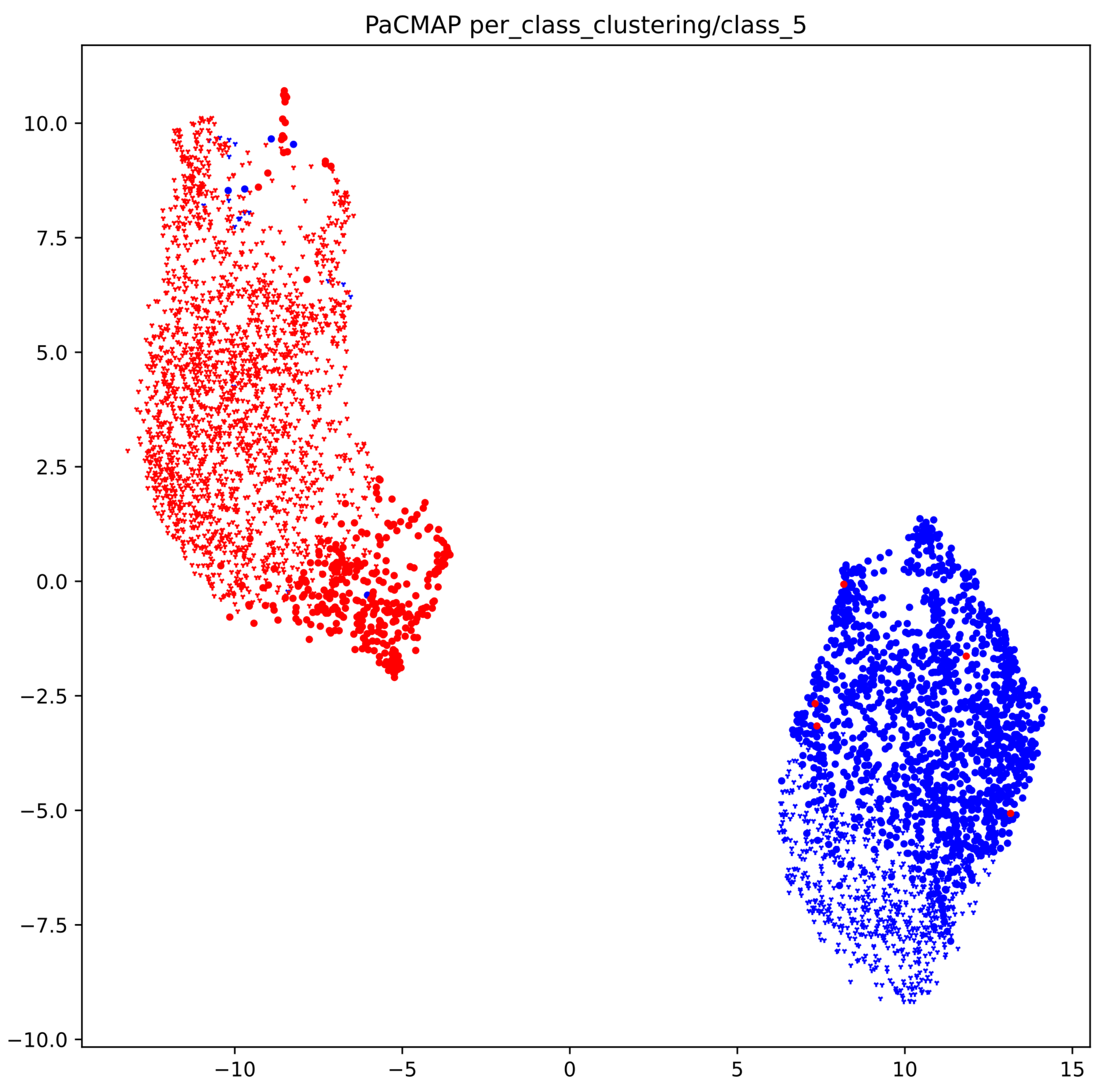}
         \caption{\textit{Class human} sample gradients (rand index 0.64)}
         \label{fig:GTA}
     \end{subfigure}
        \caption{PaCMAP representations of sample gradients of a model trained with FedAvg on Cityscapes+GTA5 for 400 rounds on the Full non-IID split. Colors represent the image domains and markers the assigned clusters by Fed-GMM. Server-side's spectral clustering is robust to imperfections of per-class clustering.}
        \label{fig:PaCMAP_city}
\end{figure}

Moreover, while each class-specific Fed-GMM gives a close to perfect domain isolation for TMNIST-Inv, they do not for Cityscapes+GTA5. We show in Figure \ref{fig:PaCMAP_city} similar PaCMAP representations for Cityscapes+GTA5 dataset. Fed-GMMs can have difficulty matching the image domains for some classes. However, linking class-specific clusterings together with server-side spectral clustering brings robustness to the final results.

\subsubsection{Pruned Deep Domain Isolation performance}

Clusterings remain perfect on TMNIST-Inv  whatever the distribution and split round. We show in Table \ref{tab:clustering_city} the clustering performance of Pruned Deep Domain Isolation with $\alpha=0.01$ on models pretrained with FedAvg on Cityscapes+GTA5. Overall, Deep Domain Isolation performs similarly to its non-pruned counterpart even when pruning 99\% of the parameters of the gradients. The domain shift is sufficiently consistent and spread over all classes to affect every part of the model, we do not need to use the full gradients for clustering.

\subsection{Test time cluster assignment evaluation}\label{sec:results_source_classif}

An average F1-score of 1.0 is obtained for domain classifiers trained on TMNIST-Inv clusterings, whatever the federated split. For Cityscapes+GTA5, they classify the image domains with an average F1-scores of 0.998, 0.997 and 0.998 when trained respectively on IID, Full non-IID and Dirichlet non-IID distributions and associated computed clustering. Even in the hardest setup of Full non-IID split with noisy labels, the domain classifier reaches close to perfect performance without overfitting on the noisy labels.

\section{Conclusion and future works}

We proposed in this work a framework for Sample Clustered Federated Learning on decentralized segmentation datasets where each client's data distribution is a different mixture of image domains. We think that this framework has a high potential, as it is much practical, and closer to the fields of Meta-Learning, Multi-Task learning or Domain Adaptation than client-based federated personalization and clustering. We showed that FedAvg on segmentation tasks simplified through Deep Domain Isolation converges faster while being agnostic to any client distribution assumption. It can also be used in a straight forward fashion on clients outside of the federation, as long as it is used on the same image domains. We only tested our framework on datasets composed of only two well distinguished domains. Further validation of the method should be led on more complex ones, although building such datasets is difficult.

We proposed in this article a baseline version of this novel framework. The choice of Fed-GMMs to cluster gradients could be refined, as the assumption of Gaussian distribution of gradients per image domain may be limiting for more complex covariate shifts. We showed in Figures \ref{fig:PaCMAP_TMNIST-Inv} and \ref{fig:PaCMAP_city} that PaCMAP is well able to capture the gradient distribution, but its application in a federated setup seems impossible for now. It would require the communication of every sample gradient for every class to the server, which seems prohibitive both in terms of communication cost and privacy. Developing advanced federated clustering algorithms for extremely high dimension spaces would improve the clustering performance and benefit our framework, as the intuition of well-behavedness of gradients per image domain in segmentation tasks seems validated by our experiments. We could also develop interactions between cluster trainings to further boost the performance instead of isolated FedAvg per cluster. We showed that gradient pruning is a good lead to reduce the dimensionality of the problem without altering sample gradient distributions. We showed that a random selection of parameters is already performing well but more specific methods could be to use curvature information for example. Other forms of test-time cluster assignment could be imagined instead of training a domain classifier. 

Finally, while we chose to focus on covariate shifts, Deep Domain Isolation should also capture concept shifts. A variant of our framework based on DDI could be applicable on ambiguous tasks such as in \cite{kohl_probabilistic_2019}. This is subject to further work.

\begin{credits}
\subsubsection{\ackname} This work was partially supported by the Agence Nationale de la Recherche under grant ANR-20-THIA-0007 (IADoc@UdL). It was granted access to the HPC resources of IDRIS under the allocation 2022-AD011013327R1 and made by GENCI.

\subsubsection{\discintname}
The authors have no competing interests to declare that are relevant to the content of this article.
\end{credits}

\bibliographystyle{splncs04}

\end{document}